\PassOptionsToPackage{capitalise}{cleveref}

\documentclass[sigconf,arxiv]{aamas}  
\AtBeginDocument{%
  \providecommand\BibTeX{{%
    \normalfont B\kern-0.5em{\scshape i\kern-0.25em b}\kern-0.8em\TeX}}}
    
\usepackage{booktabs}    
\usepackage{flushend} 

\setcopyright{ifaamas}  
\copyrightyear{2020} 
\acmYear{2020} 
\acmDOI{} 
\acmPrice{} 
\acmISBN{} 

\usepackage{nicefrac}     
\usepackage{mathtools}
\usepackage{bm}
\usepackage{thmtools}
\usepackage{thm-restate}
\usepackage{comment}
\usepackage{multirow}
\usepackage{amsmath}
\usepackage{amsfonts}
\usepackage{amssymb}
\usepackage{amsthm}
\usepackage{a4wide}
\usepackage{wrapfig}

\usepackage{mathtools}

\usepackage{appendix}
\usepackage[english]{babel}
\usepackage{graphicx}
\usepackage{subcaption}
\usepackage{xspace}
\usepackage{algorithm}
\usepackage[noend]{algpseudocode}
\usepackage{algorithmicx}
\algnewcommand\AND{\textbf{\ and\ }}
\algnewcommand{\algorithmicor}{\textbf{ or }}
\algnewcommand{\OR}{\algorithmicor}

\usepackage{bm}
\usepackage[percent,rel]{overpic}
\usepackage[export]{adjustbox}

\usepackage{relsize}
\DeclareMathOperator*{\tsum}{\mathlarger{\textstyle\sum}}

\newcommand{\rulesep}{\unskip\ \vrule\ }

\newcommand{\beq}{\begin{equation}}
\newcommand{\eeq}{\end{equation}}

\newcommand{\beqa}{\begin{eqnarray}}
\newcommand{\eeqa}{\end{eqnarray}}

\newcommand{\beqan}{\begin{eqnarray*}}
\newcommand{\eeqan}{\end{eqnarray*}}

\newcommand{\bi}{\begin{itemize}}
\newcommand{\ei}{\end{itemize}}
\newcommand{\be}{\begin{enumerate}}
\newcommand{\ee}{\end{enumerate}}

\newcommand{\E}{\mathbb{E}}

\newcommand{\NashConv}{\textsc{NashConv}\xspace}

\newcommand{\indic}[1]{\mathbb{I}\{#1\}}

\newlength{\minipagewidth}
\let\R\undefined
\newcommand{\R}{\mathbb{R}}
\newcommand{\Real}{\mathbb{R}}

\theoremstyle{acmplain}
\newtheorem*{remark*}{Remark}
\newtheorem{statement}{Statement}%
\newtheorem*{thm*}{Theorem}%

\renewcommand{\phi}{\varphi}
\renewcommand{\epsilon}{\varepsilon}

\newcommand{\A}{\mathcal{A}}

\renewcommand{\S}{\mathcal{S}}

\newcommand{\ba}{\mathbf{a}}

\newcommand{\btheta}{\bm{\theta}}
\newcommand{\bw}{\bm{w}}

\newcommand{\bpi}{\bm{\pi}}

\newcommand{\abs}[1]{\left|#1\right|}

\DeclareMathOperator*{\argmax}{arg\,max}

\DeclarePairedDelimiter{\norm}{\lVert}{\rVert}

\newcommand{\States}{\S}

\usepackage{dsfont}
\newcommand{\as}{\doteq}
\newcommand{\subex}[1]{\left(#1\right)}
\newcommand{\subblock}[1]{\left[#1\right]}
\newcommand{\set}[1]{\left\{ {#1} \right\}}

\newcommand{\reals}{\Real}
\newcommand{\simplex}{\varDelta}

\newcommand{\littleO}[1]{\operatorname{o}\subex{#1}}

\newcommand{\indicator}[1]{\mathds{1}_{#1}}
\newcommand{\e}[1]{e^{#1}}
\newcommand{\policy}{\pi}

\newcommand{\Actions}{\A}
\newcommand{\action}{a}
\newcommand{\numActions}{\abs{\Actions}}

\newcommand{\projection}{\Pi}
\newcommand{\softmaxProjection}{\projection}

\newcommand{\ones}{\bm{1}}

\newcommand{\utility}{u}

\newcommand{\learningRate}{\eta}
\newcommand{\stepSize}{\learningRate}

\newcommand{\Regret}{R}

\newcommand{\logit}{y}

\newcommand{\bec}{\bm}
\newcommand{\logits}{\bec{\logit}}
\newcommand{\jointUtility}{\bec{\utility}}
\newcommand{\grad}{\nabla}

\newcommand{\heads}{$\mathcal{H}$\xspace}
\newcommand{\tails}{$\mathcal{T}$\xspace}
\newcommand{\even}{``even''\xspace}
\newcommand{\odd}{``odd''\xspace}

\newcommand{\defword}{\emph}

\newcommand{\neurdfull}{Neural Replicator Dynamics\xspace}
\newcommand{\neurd}{NeuRD\xspace}

\renewcommand\cite{\GenericError{}
    {Error: \string\cite\space should not be used!}
    {The standard LaTeX command `\string\cite' should be avoided, because it behaves like `\string\citet' for author-year citations, but like `\string\citep' for numerical ones.}{}}%

\usepackage{autonum}

\usepackage{chngcntr}

\crefname{equation}{}{}
\crefname{appsec}{Appendix}{Appendices}

\crefname{paragraph}{Section}{Sections}
\Crefname{paragraph}{Section}{Sections}

\newcounter{apcounter}
\newcommand{\appendixproof}[1]{
    \par
    \refstepcounter{apcounter}
    \textbf{\theapcounter}\space\space
    \textbf{#1}
    }
\crefname{apcounter}{Section}{Sections}
\Crefname{apcounter}{Section}{Sections}
\crefformat{apcounter}{#2Section~A.#1#3}
\Crefformat{apcounter}{#2Section~A.#1#3}

\newif\ifarxiv

\arxivtrue 

\ifarxiv
    \acmConference[]{}{}{}
\fi

\begin{document}

\title{Neural Replicator Dynamics:\\ Multiagent Learning via Hedging Policy Gradients}
\subtitle{}

\author{Daniel Hennes}
\authornote{Equal contributors.}
\authornote{DeepMind.}
\affiliation{}
\email{hennes@google.com}

\author{Dustin Morrill}
\authornotemark[1]
\authornote{Work done during an internship at DeepMind.}
\affiliation{}
\email{morrill@ualberta.ca}

\author{\mbox{Shayegan~Omidshafiei}}
\authornotemark[1]
\authornotemark[2]
\affiliation{}
\email{somidshafiei@google.com}

\author{R\'emi Munos}
\authornotemark[2]
\affiliation{}
\email{munos@google.com}

\author{Julien Perolat}
\authornotemark[2]
\affiliation{}
\email{perolat@google.com}

\author{Marc Lanctot}
\authornotemark[2]
\affiliation{}
\email{lanctot@google.com}

\author{Audrunas Gruslys}
\authornotemark[2]
\affiliation{}
\email{audrunas@google.com}

\author{Jean-Baptiste Lespiau}
\authornotemark[2]
\affiliation{}
\email{jblespiau@google.com}

\author{Paavo Parmas}
\authornotemark[3]
\affiliation{}
\email{paavo.parmas@oist.jp}

\author{\mbox{Edgar~Du\'e\~nez-Guzm\'an}}
\authornotemark[2]
\affiliation{}
\email{duenez@google.com}

\author{Karl Tuyls}
\authornotemark[2]
\affiliation{}
\email{karltuyls@google.com}

\renewcommand{\shortauthors}{D.~Hennes et al.}

\begin{abstract}
Policy gradient and actor-critic algorithms form the basis of many commonly used training techniques in deep reinforcement learning. 
Using these algorithms in multiagent environments poses problems such as nonstationarity and instability.
In this paper, we first demonstrate that standard softmax-based policy gradient can be prone to poor performance in the presence of even the most benign nonstationarity.
By contrast, it is known that the replicator dynamics, a well-studied model from evolutionary game theory, 
eliminates dominated strategies and exhibits convergence of the time-averaged trajectories to interior Nash equilibria in zero-sum games.
Thus, using the replicator dynamics as a foundation, we derive an elegant one-line change to policy gradient methods that simply bypasses the gradient step through the softmax, yielding a new algorithm titled \defword{\neurdfull (\neurd)}.
\neurd reduces to the exponential weights/Hedge algorithm in the single-state all-actions case.
Additionally, \neurd has formal equivalence to softmax counterfactual regret minimization, which guarantees convergence in the sequential tabular case. Importantly, our algorithm provides a straightforward way of extending the replicator dynamics to the function approximation setting.
Empirical results show that \neurd quickly adapts to nonstationarities, outperforming policy gradient significantly in both tabular and function approximation settings, when evaluated on the standard imperfect information benchmarks of Kuhn Poker, Leduc Poker, and Goofspiel.
\end{abstract}


\maketitle

\section{Introduction}

Policy gradient (PG) algorithms form the foundation of many scalable approaches driving the field of deep reinforcement learning (RL)~\citep{schulman2015trust,mnih2016asynchronous,lillicrap2015continuous,espeholt2018impala,schulman2017proximal}.
Agents using PG-based algorithms have learned to navigate in complex 3D worlds, play a wide array of video games, and learned to simulate
humanoid locomotion on high-dimensional continuous control problems.
The problem of \defword{multiagent reinforcement learning} (MARL)~\citep{BusoniuBS08,PanaitL05,TuylsW12}, which involves several agents acting and learning simultaneously, is significantly more challenging because each agent perceives its environment as nonstationary~\citep{matignon2012independent,TuylsW12}.
There have been several extensions or applications of PG algorithms to the multiagent setting, with remarkable success~\citep{lowe2017multi,foerster2018counterfactual,foerster2018learning,Bansal17Emergent,AlShedivat18Continuous}.
Given the wide use of PG and related variants in practice, it is paramount to understand its behavior and potential failure modes.

In partially-observable zero-sum games, such as poker, traditional approaches using expert knowledge or search with a perfect
model have scaled to very large games~\citep{bowling2015heads,Moravcik17DeepStack,Brown17Libratus,Brown19Pluribus}.
However, there have been RL-inspired advances as well:
for example Regression Counterfactual Regret Minimization (RCFR)~\citep{Waugh15,dorazio2019alternative}, Neural Fictitious Self-Play (NFSP)~\citep{heinrich2016deep}, Policy-Space Response Oracles (PSRO)~\citep{Lanctot17}, 
Deep Counterfactual Regret Minimization~\citep{Brown18DeepCFR}, Double Neural CFR~\citep{Li20DNCFR}, and Exploitability Descent (ED)~\citep{ED-arXiv}.
Recent work proposed Regret PGs (RPG)~\citep{srinivasan2018actor},
an entirely model-free RL approach with formal relationships established between action-values and counterfactual values used in tabular regret minimization algorithms for partially observable zero-sum games.
With a tabular representation, we can achieve regret minimization guarantees with the \defword{Hedge} algorithm~\citep{Freund97,Littlestone94}\footnote{This algorithm has many names including exponential weights, multiplicative weights, and entropic mirror descent~\citep{beck2003mirror}.} or its bandit version, \defword{Exp3}~\citep{auer2002exp3}, but they are not trivially extended to the function approximation case.

\defword{Evolutionary game theory} (EGT) studies how populations of individuals interact strategically~\citep{Smith73,Weibull97,Hofbauer98}.
EGT requires minimal knowledge of opponent strategies or preferences, and has been important for the analysis and evaluation of MARL agents~\citep{tuyls2004evolutionary,PonsenTKR09,TuylsSym,TuylsPLLG18,omidshafiei2019alpha}.
Formal connections have been made between MARL and EGT~\citep{Tuyls03,BloembergenTHK15}, and  population-based training regimes have been critical in scaling MARL to complex multiagent environments~\citep{JaderbergCTF,Bard18Hanabi,Vinyals19AlphaStar}.
Central to EGT is the standard \defword{replicator dynamics}~(RD): a dynamical system that increases/decreases the tendency of
the population toward playing strategies that would give high/low payoffs, compared to the average population value.

In this paper, we introduce a novel algorithm, \defword{\neurdfull}~(\neurd), that corresponds to a parameterization of RD.
Specifically, \neurd is a fully sample-based and model-free algorithm for sequential nonstationary environments which is not restricted to the case of perfect information Markov/stochastic (simultaneous move) games~\citep{Hennes09State}, does not require sequence-form representations~\citep{Panozzo14SFQ}, and is fully compatible with general function approximation (e.g., deep neural networks).
The new algorithm is an elegant one-line modification of the standard softmax PG, which effectively skips the gradient through the final softmax layer. 
An important result of this transformation is the ability to respond more dynamically to environmental nonstationarities.
\neurd is provably no-regret, which yields time-average convergence to approximate Nash equilibria in partially-observable zero-sum games under standard softmax policy representations. Our results validate these claims empirically in several partially observable games, even with compounded nonstationarity in the environment in the form of a dynamically-changing reward function.

\subsection{Motivating Example}\label{sec:motivating_example}
Softmax Policy Gradient (SPG) is usually deployed in complex domains, implemented with function approximation, and trained via stochastic gradients. 
Here we detail a simple case where SPG fails to be reasonably adaptive even with these complexities removed.

Consider \defword{matching pennies}, a two-player zero-sum game with two actions, \heads and \tails. If both players choose the same action, the \even player receives a reward of 1 and the \odd player receives a reward of -1, and vice-versa if the players choose opposing actions. 
We will simplify the problem further by considering only the policy for the \even player and fixing the \odd player's policy in each playout:
given a number of rounds to play in advance, the \odd player will play \heads for the first 40\% of the rounds and play \tails for the remaining 60\%. 
We will consider the all-actions version of the repeated game, where players are evaluated by their cumulative expected reward and observe their entire reward functions.

What would be a good score for the \even player to achieve?
A starting point is to look at the value of fixed pure strategies. Playing \heads over $T$-rounds achieves a value of $-0.2T$, while playing \tails achieves $0.2T$. 
The difference between the player's value and the cumulative reward of the best fixed action in hindsight is called \defword{regret}, and ensuring that this difference does not increase with the horizon length is evidence that an algorithm can robustly learn to distinguish between good and bad actions in a reasonable time. 

Let us now compare SPG to Hedge~\citep{Freund97,Littlestone94}, a well-known no-regret algorithm. 
While Hedge has near-constant regret\footnote{In fact, we can ensure that Hedge achieves at most a constant regret of 1 for any horizon length $T$ as the \even player by simply choosing a large step size.},  SPG's regret grows with the horizon length, even if the step size is independently tuned for each $T$.
The growth rate is roughly a function of $\sqrt{T}$ though, which matches the order of the best worst-case performance bound.

\begin{figure}[t]
    \centering
    \includegraphics[width=0.9\linewidth]{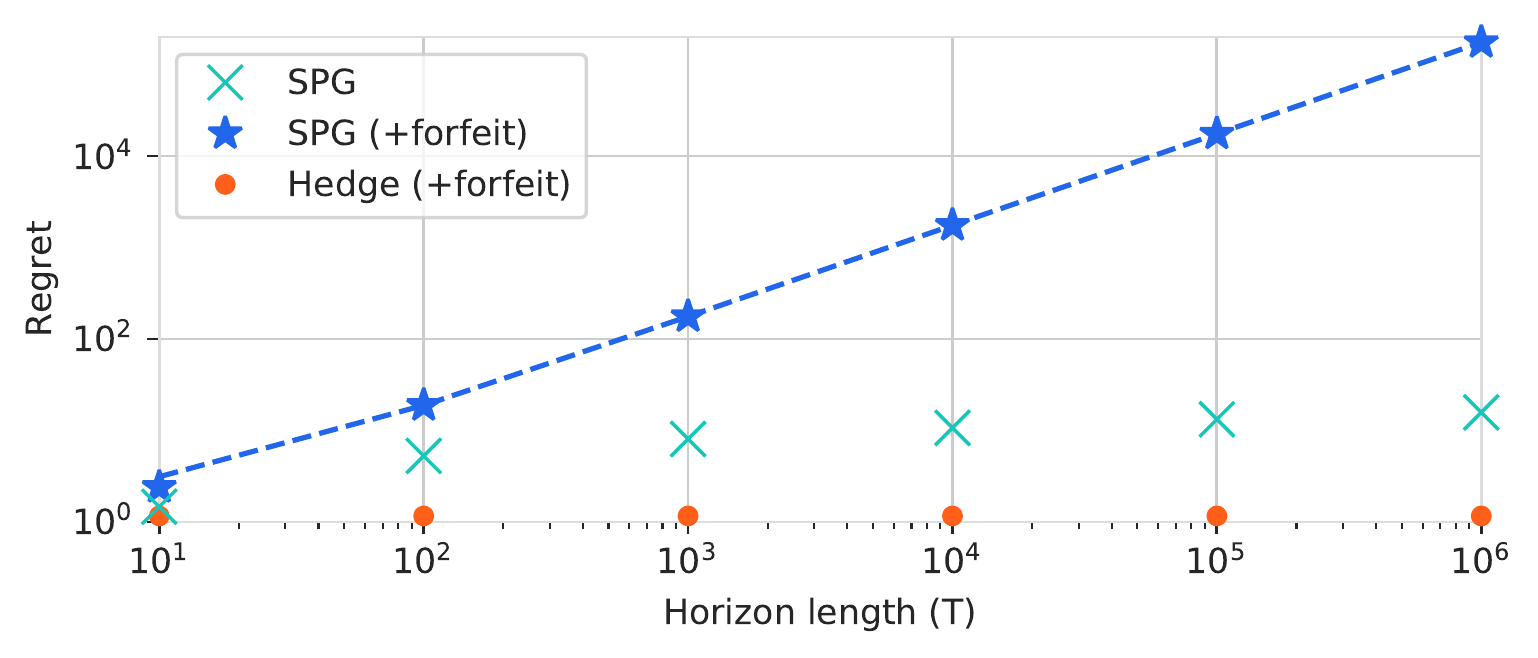}
    \vspace{-10pt}
    \caption{The regret of SPG with and without a forfeit action in repeated matching pennies compared to Hedge. The dashed line is a linear least-squares fit.}
    \label{fig:threeActionEnv}
\end{figure}

However, the difference is amplified, if we simply add a nuisance action choice for the \even player. This action only has the effect of forfeiting the round to the \odd player, so its reward is always -1. \Cref{fig:threeActionEnv} shows that this drastically increases the regret growth-rate for SPG to linear, which is larger than worst-case bounds provided by other algorithms (including Hedge) by a factor of $\sqrt{T}$. In practice, Hedge's regret remains \emph{constant}.
This example, trivially, shows that even in the simplest of non-stationary environments SPG does not respond to changes in a reasonable manner, while other simple algorithms perform well.
We will revisit this example in ~\cref{sec:pg_limitations} and further explore why SPG's regret grows so rapidly.%

\section{Preliminaries}\label{sec:preliminaries}
We first establish prerequisite definitions and notation before presenting our technical analysis of SPG and the proposed algorithm.

\subsection{Game Theory}\label{sec:game_theory} 
Game theory studies strategic interactions of players.
A \defword{normal-form game} (NFG) specifies the interaction of $K$ players with corresponding action sets $\{\Actions^{1}, \ldots, \Actions^{K}\}$. 
The payoff function $\jointUtility: \prod_{k=1}^{K} \Actions^{k} \mapsto \R^K$ assigns a numerical utility to each player for each possible joint action $\ba \as (\action^{1}, \ldots, \action^{K})$, where $\action^{k} \in \Actions^{k}$ for all $k \in [K] \as \{1,\ldots,K\}$.
Let $\bpi^{k} \in \simplex^{\numActions}$ denote the $k$-th player's mixed strategy. 
The expected utility for player $k$ given strategy profile $\bpi \as (\bpi^{1}, \ldots, \bpi^{K})$ is then $\bar{u}^{k}(\bpi) \as \E_{\bpi}[\utility^{k}(\ba) | \ba \sim \bpi]$.
The best response for player $k$ given $\bpi$ is $\text{BR}^{k}(\bpi^{-k}) = \argmax_{\bpi^{k}}  [\bar{u}^{k}(({\bpi^{k}, \allowbreak \bpi^{-k}}))]$, where $\bpi^{-k}$ is the set of opponent policies.
Profile $\bpi_{*}$ is a Nash equilibrium if $\bpi_{*}^{k} = \text{BR}^{k}(\bpi_{*}^{-k})$ for all $k \in [K]$.
We use the Nash Convergence metric (\NashConv)~\citep{Lanctot17} to evaluate learned policies:
\begin{equation}
    \NashConv(\bpi) = \tsum_{k}  \bar{u}^{k}(({\text{BR}^{k}(\bpi^{-k}),\bpi^{-k}})) - \bar{u}^{k}({\bpi})
    \label{eq:nashconv}
\end{equation}
Roughly speaking, \NashConv measures `distance' of $\bpi$ to a Nash equilibrium (i.e., lower \NashConv is better).

\subsection{Replicator Dynamics (RD)}
Replicator Dynamics (RD) is a concept from EGT that describe a population's evolution via biologically-inspired operators, such as selection and mutation~\citep{Taylor78,Taylor79,Hofbauer98,Zeeman80,Zeeman81}. 
The single-population RD are defined by the following system of differential equations:
\begin{equation}
    \dot{\pi}(a) = \pi(a) \big[ u(a, \bpi) - \bar{u}(\bpi)\big] \qquad \forall a \in \Actions,
    \label{eq:rd}
\end{equation}
Each component of $\bpi$ determines the proportion of an action $a$ being played in the population.%
The time derivative of each component is proportional to the difference in its expected payoff, $u(a,\bpi)$, and the average payoff, $\bar{u}(\bpi) = \sum_{a\in\Actions} \pi(a) u(a,\bpi)$.

\subsection{Online Learning}
\label{sec:online-learning}

Online learning examines the performance of learning algorithms in potentially adversarial environments. 
On each round, $t$, the learner samples an action, $\action_t \in \Actions$ from a discrete set of actions, $\Actions$, according to a policy, $\bpi_t \in \simplex^{\numActions}$, and receives utility, $\utility_t(\action) \in \reals$, where $\bm{\utility}_t \in \reals^{\numActions}$ is a bounded vector provided by the environment. A typical objective is for the learner to minimize its expected \emph{regret} in hindsight for not committing to $\action \in \Actions$ after observing $T$ rounds; the regret is defined as $\Regret_T(\action) \as \sum_{t=1}^T \utility_t(\action)  - \bpi_t \cdot \bm{\utility}_t$.
Algorithms that guarantee their average worst-case regret goes to zero as the number of rounds increases, i.e., $\Regret_T \in \littleO{T}$, are called \emph{no-regret}; 
these algorithms learn optimal policies under fixed or stochastic environments. According to a folk theorem, the average policies of no-regret algorithms in self-play or against best responses converge to a Nash equilibrium in two-player zero-sum games~\citep{Blum07}. This result can be extended to sequential imperfect information games by composing learners in a tree and defining utility as counterfactual value~\citep{Zinkevich08,hofbauer2009time}.
We refer the interested reader to \citep{srinivasan2018actor} for additional background on sequential games.

The family of no-regret algorithms known as Follow the Regularized Leader (FoReL)~\citep{mcmahan2011follow,shalev2007online,shalev2007primal,mcmahan2013ad} generalizes well-known decision making algorithms and population dynamics. 
For a discrete action set, $\Actions$, FoReL is defined through the following updates:
\begin{align}
    \bm{\policy}_t &\as \argmax_{\bm{\policy}' \in \simplex^{\numActions}} \subblock{  \bm{\policy}' \cdot \bm{\logit}_{t - 1} - h(\bm{\policy}') },\label{eq:forel_discrete_policy} \; \qquad
    \bm{\logit}_t \as \bm{\logit}_{t - 1} + \stepSize_t \bm{u}_t,
\end{align}
where $\learningRate_t > 0$ is the learning rate at timestep $t$, $\bm{\utility}_t \in \reals^{\numActions}$ is the utilities vector observed at $t$, $\bm{\logit}_t \in \reals^{\numActions}$ are the accumulated values at $t$, and regularizer $h$ is a convex function.
Note that FoReL assumes that the learner observes the entire action utility vector at each timestep, rather than only the reward for taking a particular action. 
This is known as the \defword{all-actions} setting.

Under negative entropy regularization $h(\bpi)=\sum_{\action} \pi(\action) \log \pi(\action)$, policy $\bpi_t$ reduces to a softmax function $\bm{\policy}_t \as \softmaxProjection(\bm{\logits}_{t - 1})$, where  $\softmaxProjection(\bm{z}) \propto \exp(z), \forall z \in \reals^{\numActions}$.
This yields the Hedge algorithm:
\begin{align}
    \bec{\policy}_T \as \softmaxProjection
        \subex{ \textstyle\tsum_{t = 1}^{T - 1} \learningRate_t \bm{\utility}_t }.
    \label{eq:hedge}
\end{align}
Hedge is no-regret when $\stepSize_t$ is chosen carefully, e.g., $\stepSize_t \in \Theta(\nicefrac{1}{\sqrt{t}})$~\citep{Nedic14}. 
Likewise, the continuous-time FoReL dynamics~\citep{mertikopoulos2018cycles} are
\begin{align}
    \bpi &\as \argmax_{\bpi' \in \simplex^{\numActions}} \big[ \bpi' \cdot \logits - h(\bpi') \big], \; \qquad
   \dot{\bm{\logit}} \as \bm{\utility}, \label{eq:forel_logits}
\end{align}
which in the case of entropy regularization yield RD as defined in~\eqref{eq:rd} (e.g., see~\citet{mertikopoulos2018cycles}).
This implies that RD is no-regret, thereby enjoying equilibration to Nash and convergence to the optimal prediction in the time-average.

\begin{figure}[t]
    \centering
    \includegraphics[width=\linewidth]{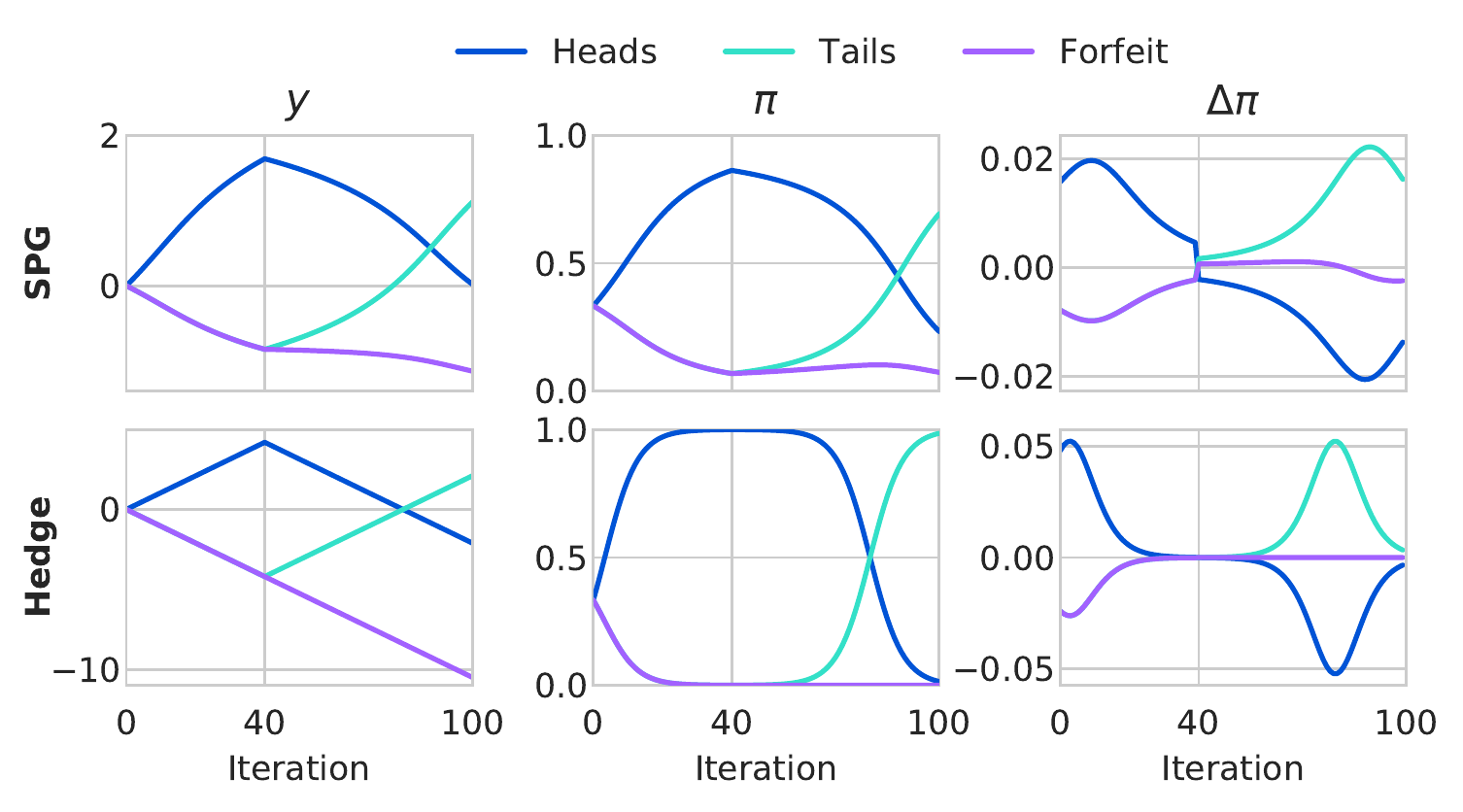}
    \vspace{-20pt}
    \caption{
The logit and policy trajectories of SPG and Hedge in all-actions, 100-round, repeated matching pennies with a forfeit action. The vertical lines mark the change in the opponent's policy at 40-rounds. 
The step size $\eta=0.21$ was optimized in a parameter sweep for SPG with $T=100$.}
    \label{fig:hedge_spg}
\end{figure}

\subsection{Policy Gradient (PG)}
In a Markov Decision Process, at each timestep $t$, an agent in state $s_t \in \States$ selects an action $a_t \in \Actions$, receives a reward $r_t \in \reals$, then transitions to a new state $s_{t+1} \sim \mathcal{T}(s,a,s')$.
In the discounted infinite-horizon regime, the reinforcement learning (RL) objective is to learn a policy $\bpi: s \to \simplex^{\numActions}$, which maximizes the expected return $v^{\bpi}(s) = \mathbb{E}_{\bpi}[\sum_{k=t}^\infty \gamma^{k-t} r_k | s_t = s]$, with discount factor $\gamma \in [0,1)$.  
In actor-critic algorithms, one generates trajectories according to some parameterized policy $\pi(\cdot|s;\btheta)$ while learning to estimate the action-value function $q^{\bpi}(s,a) = \mathbb{E}_{\bpi}[\sum_{k=t}^\infty \gamma^{k-t} r_k | s_t = s, a_t = a]$. 
Temporal difference learning~\citep{Sutton18} can be used to learn an action-value function estimator, $q(s,a;\bec{w})$, which is parameterized by $\bec{w}$.
A PG algorithm then updates policy $\bpi$ parameters $\btheta$ in the direction of the gradient $\nabla_{\btheta}\log\pi(a|s;\btheta) \big[q(s,a;\bec{w}) - v(s;\bec{w})\big]$ for a given state-action pair $(s,a)$, where the quantity in square brackets is defined as the \emph{advantage}, denoted $A(a;\btheta,\bec{w})$, and $v(s;\bec{w}) \as \sum_{a'}\pi(s,a';\btheta)q(s,a';\bec{w})$. 
The advantage is analogous to regret in the online learning literature. 
In sample-based learning, the PG update incorporates a $(\policy(\action | s; \btheta))^{-1}$ factor that accounts for the fact that $\action$ was sampled from $\bpi$. 
The all-actions PG update without this factor is then 
\begin{align}
    \btheta_t = \btheta_{t-1} + \eta_t\tsum_{a}\nabla_{\bec{\btheta}}\policy( a | s ; \btheta_{t-1}) \big[{q}(s, a ; \bec{w}) - v(s; \bec{w}) \big].
    \label{eq:pg_discrete}
\end{align}
While different policy parameterizations are possible, the most common choice for discrete decision problems is a {softmax function} over the \defword{logits} $\logits$: 
$\bpi_{t}(\btheta_{t}) = \softmaxProjection(\logits(\btheta_{t}))$.
Thus, we focus the rest of our analysis on Softmax Policy Gradient (SPG).

\section{A Unifying Perspective on RD and PG}\label{sec:unifying}
This section motivates and presents a novel algorithm, \neurdfull (\neurd), and unifying theoretical results.

\begin{figure*}[!t]
    \hspace{-2em}
    \begin{subfigure}[t]{0.195\linewidth}
        \begin{overpic}[width=\textwidth,trim={2mm 2mm 4mm 2mm},clip]{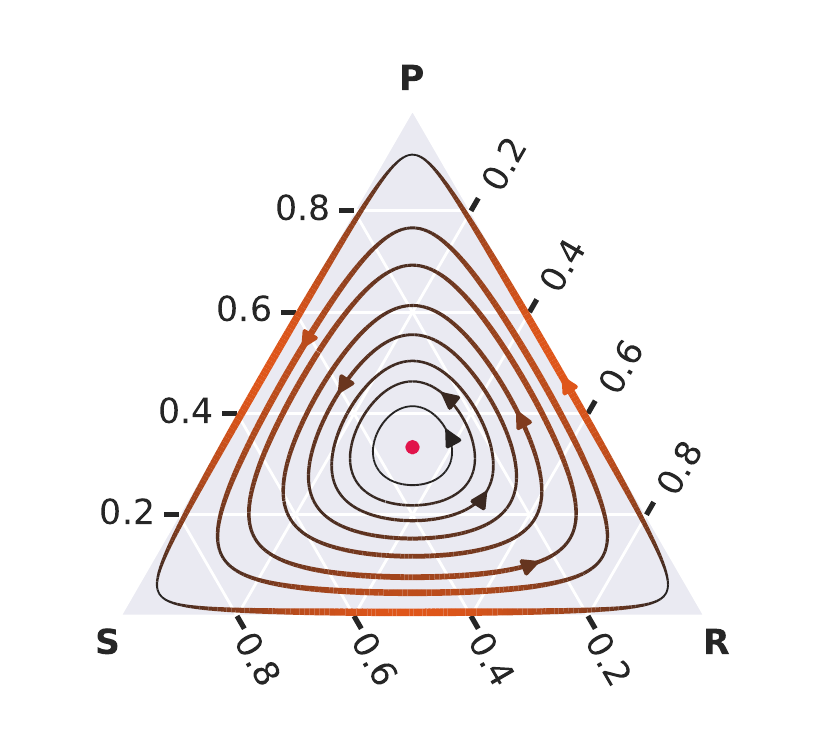}
            \put(5,80){\small\textbf{(a)}}
        \end{overpic}
        \phantomcaption
        \label{fig:rps_rd}
    \end{subfigure}%
    \begin{subfigure}[t]{0.195\linewidth}
        \begin{overpic}[width=\textwidth,trim={2mm 2mm 4mm 2mm},clip]{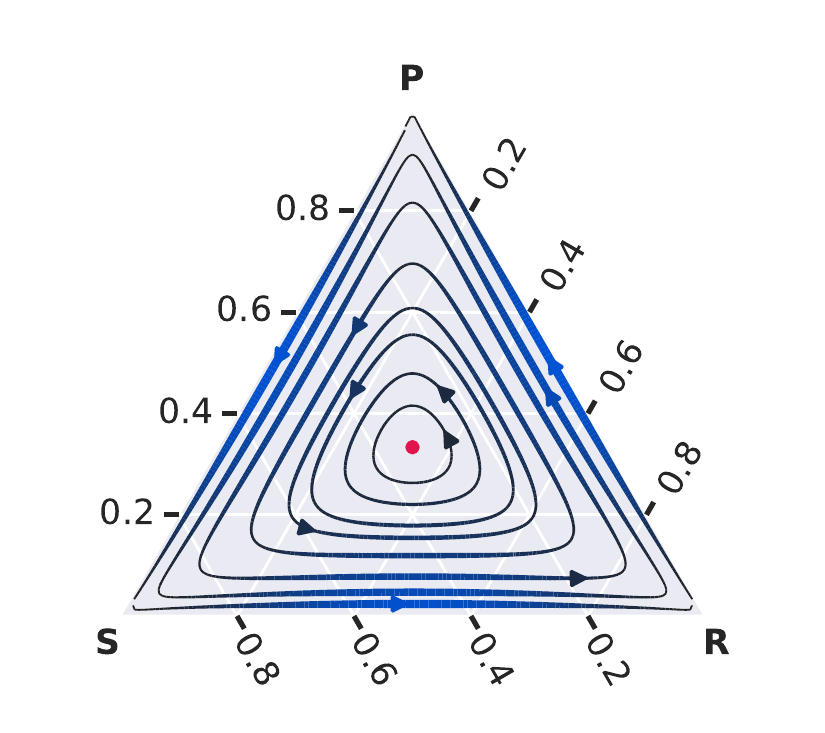}
            \put(5,80){\small\textbf{(b)}}
        \end{overpic}
        \phantomcaption
        \label{fig:rps_qpg}
    \end{subfigure}
    \begin{subfigure}[t]{0.195\linewidth}
        \begin{overpic}[width=\textwidth,trim={2mm 2mm 4mm 2mm},clip]{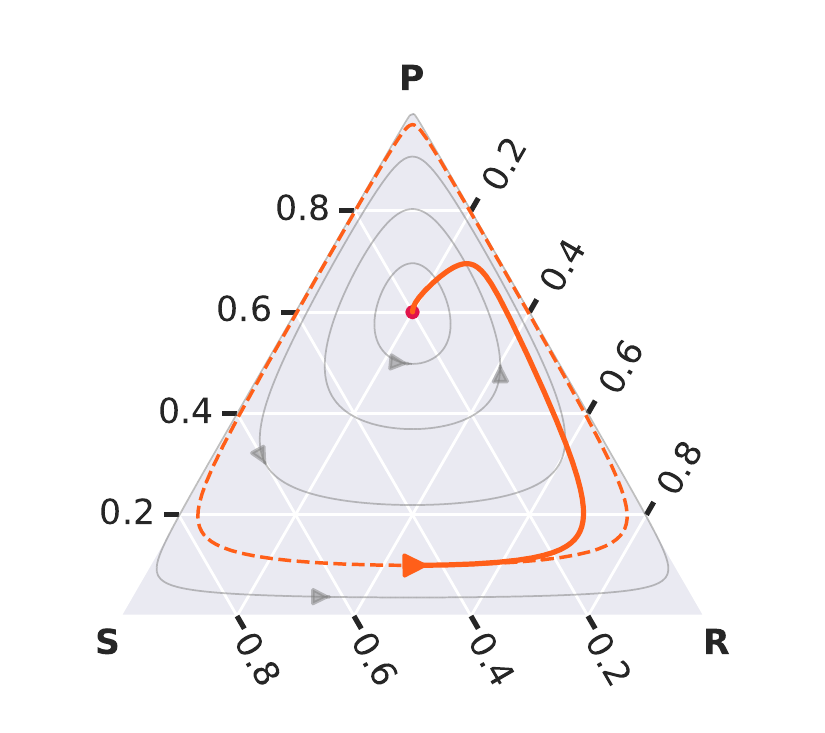}
            \put(5,80){\small\textbf{(c)}}
        \end{overpic}
        \phantomcaption
        \label{fig:biased_rps_rd}
    \end{subfigure}
    \begin{subfigure}[t]{0.195\linewidth}
        \begin{overpic}[width=\textwidth,trim={2mm 2mm 4mm 2mm},clip]{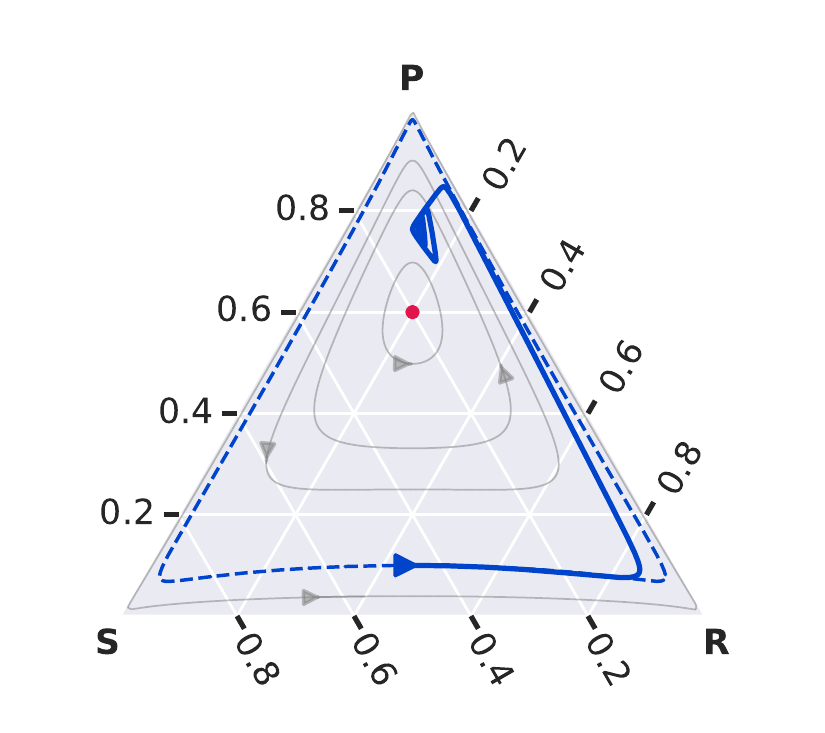}
            \put(5,80){\small\textbf{(d)}}
        \end{overpic}
        \phantomcaption
        \label{fig:biased_rps_qpg}
    \end{subfigure}
    \begin{subfigure}[t]{0.195\linewidth}
        \begin{overpic}[width=\textwidth,trim={2mm 2mm 4mm 2mm},clip]{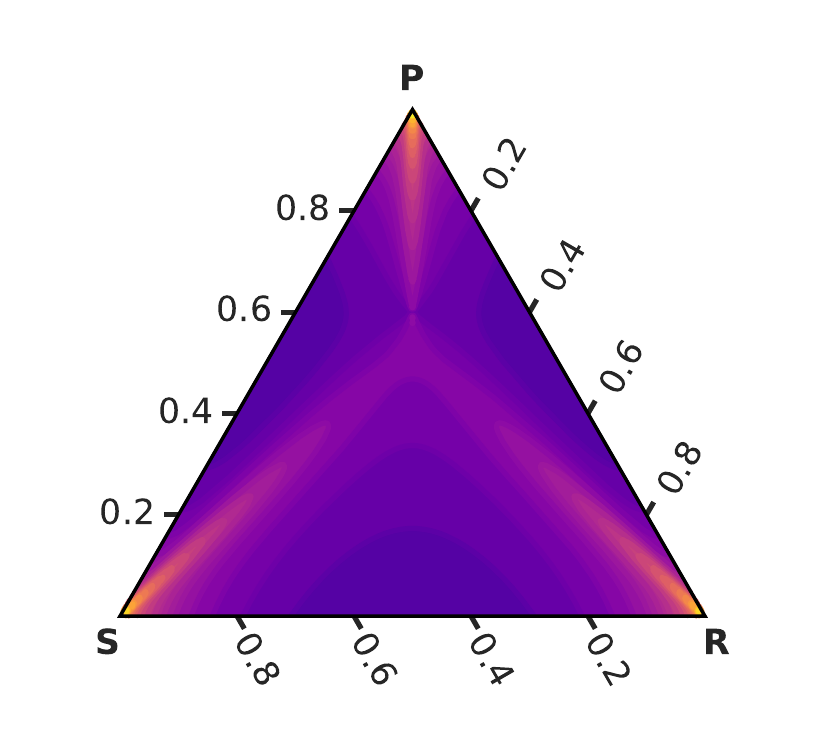}
            \put(5,80){\small\textbf{(e)}}
            \put(100,10){\includegraphics[height=.7\textwidth]{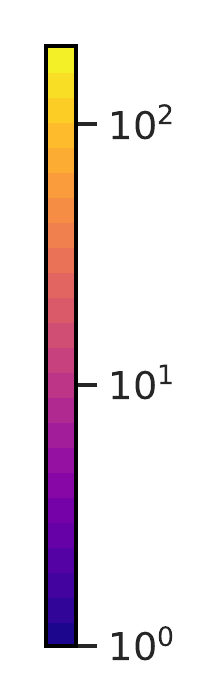}}
        \end{overpic}
        \phantomcaption
        \label{fig:rps_ratio}
    \end{subfigure}
    \vspace{-2em}
    \caption{Learning dynamics of \subref{fig:rps_rd} RD and \subref{fig:rps_qpg} SPG in \emph{Rock--Paper-Scissors} (RPS). Time-averaged trajectories (solid lines) are shown in \subref{fig:biased_rps_rd} for RD and in \subref{fig:biased_rps_qpg} for SPG in the biased-RPS game. In \subref{fig:rps_ratio} we compare their rate of adaptation, i.e., $\norm{\dot{\bm{\pi}}_\text{RD}} / \norm{\dot{\bm{\pi}}_\text{PG}}$.}
    \label{fig:rps_dynamics}\vspace{-.5em}
\end{figure*}

\subsection{A Close-up on Learning Dynamics}\label{sec:pg_limitations}
Let us consider the strengths and weaknesses of the algorithms described thus far.
While RD and the closely-related FoReL are no-regret and enable learning of equilibria in games, they are limited in application to tabular settings.
By contrast, SPG is applicable to high-dimensional single and multiagent RL domains.
Unfortunately, SPG suffers from the fact that increasing the probability of taking an action that already has low probability mass can be very slow, in contrast to the considered no-regret algorithms.
We can see this by writing out the single state, tabular, all-actions SPG update explicitly, using the notation of online learning to identify correspondences to that literature.
On round $t$, SPG updates its logits and policy as
\begin{align}
    \bec{\policy}_{t}
        &\as \softmaxProjection(\bec{\logit}_{t - 1}), \; \qquad
    \bec{\logit}_{t} 
        \as \bec{\logit}_{t - 1} + \stepSize_t \grad_{\bec{\logit}_{t - 1}} \bec{\policy}_{t}\cdot\bec{\utility}_t.
\end{align}
As there is no action or state sampling in this setting, shifting all the payoffs by the expected value $\bar{u}$ (or $v^{\bpi}$ in RL terms) has no impact on the policy, so this term is omitted above. 
Noting that $\nicefrac{\partial \policy_t(\action')}{\partial \logit_{t - 1}(\action)} = \policy_t(\action') \subblock{ \indicator{\action' = \action} - \policy_t(\action) }$~\citep[Section 2.8]{Sutton18}, we observe that the update direction, $\grad_{\bec{\logit}_{t - 1}} \bec{\policy}_t \cdot \bec{\utility}_t$, is actually the instantaneous regret scaled by $\bec{\policy}_t$, yielding the concrete update:
\begin{align}
    \logit_t(\action) 
        &= \logit_{t - 1}(\action) + \stepSize_t \policy_t(\action) \big[
        \utility_t(\action)
            - \bar{u}_t
        \big] \quad \forall a \in \Actions
    \label{eq:tabular_all_actions_pg_update}.
\end{align}
See~\cref{sec:tabular_pg_update} for details.
Scaling the regret by $\bpi_t$ leads to an update that can prevent SPG from achieving reasonable performance.

Specifically, the additional scaling factor $\bpi_t$ can hinder learning in nonstationary settings (e.g., in games) when an action might be safely disregarded at first, but later the value of this action improves.
This is precisely why SPG fails to adapt in our motivating example (\cref{sec:motivating_example}):
at the switch point $t=0.4T$, the rate of increase of the logit corresponding to \tails is modulated by a low action probability. 
Likewise, the rate of decrease of the logit corresponding to \heads is modulated by a low instantaneous regret, i.e., $\bar{u}_{t}$ is close to -1 as the probability of playing \heads is close to 1 at the switching point. 
Thus, we clearly see a difference in the slopes of the logit trajectories for SPG and Hedge in~\cref{fig:hedge_spg}, where SPG reacts slower. 
We also observe that the logit for the forfeit action is decreasing very slowly after the switch, as both the action probability and the instantaneous regret are small. 
This decreases the logit gap (i.e., the range between largest and smallest logit), thereby effectively increases probability mass on the undesirable forfeit action. 
Without an nuisance action, the logit gap would quickly reduce which in turn would reduce the probability mass on \heads due to the softmax projection. With the added asymmetry caused by the presence of a forfeit action, the logit gap remains larger, thus leaving more probability mass on \heads.
Notice that this is in stark contrast to the behavior of Hedge, where the logit for the forfeit action keeps decreasing, thus maintaining a constant logit gap until \tails becomes the preferred action.

The scaling by $\bec{\pi}_t$ is also apparent when taking the continuous-time limit of SPG dynamics. 
Consider the continuous-time $q$-value based policy gradient dynamics~\citep[Section D.1.1, QPG]{srinivasan2018actor}, which are amenable for comparison against RD:
\begin{align}
    \dot{\pi}(a;\btheta) = \pi(a;\btheta) \Big( \pi(a;\btheta) A(a,\btheta,\bw) - \tsum_b \pi(b;\btheta)^2 A(b,\btheta,\bw) \Big),
    \label{eq:qpg}
\end{align}
conducted for all $a \in \Actions$.
In contrast to RD \cref{eq:rd}, the SPG dynamics in \cref{eq:qpg} have an additional $\pi(a;\btheta)$ term that modulates learning.

The issue arising due to regret scaling manifests itself also when considering convergence to Nash equilibria in games, even under continuous-time dynamics.
We illustrate this in the game of \defword{Rock--Paper--Scissors}, by respectively comparing the continuous-time dynamics of RD and SPG in \cref{fig:rps_rd,fig:rps_qpg}.
The game is intransitive, thus resulting in trajectories that cycle around the Nash equilibrium.
In \cref{fig:biased_rps_rd} and \cref{fig:biased_rps_qpg}, we show time-averaged trajectories in a biased version of RPS (see \cref{eq:biased-rps} with $\nu=3$).
The time-averaged trajectories of RD converge to interior Nash equilibria in zero-sum games~\citet{Hofbauer09Time}.
However, even in this simple game, SPG does not converge to Nash in the time-average.
As a further comparison, \cref{fig:rps_ratio} plots the ratio of their speeds,  i.e., $\norm{\dot{\bm{\pi}}_\text{RD}} / \norm{\dot{\bm{\pi}}_\text{PG}}$ in the biased version of the game.
The differences in updates are compounded near the simplex vertices, where a single action retains a majority of policy mass.
This difference causes practical issues when using SPG in settings where learning has converged to a near-deterministic policy and then must adapt to a different policy given (e.g., under dynamic payoffs or opponents).
While SPG fails to adapt rapidly to the game at hand due to its extra downscaling by factor $\bpi$, RD does not exhibit this issue.

Given these insights, our objective is to derive an algorithm that combines the best of both worlds, in that it is theoretically-grounded and adaptive in the manner of RD, while still enjoying the practical benefits of the parametric SPG update rule in RL applications. 

\subsection{\neurd: \neurdfull}

While we have highlighted key limitations of SPG in comparison to RD, the latter has limited practicality when computational updates are inherently discrete-time or a parameterized policy is useful for generalizability.
To address these limitations, we derive a discrete-time parameterized policy update rule, titled \neurdfull (\neurd), which is later compared against SPG.
For seamless comparison of our update rule to SPG, we next switch our nomenclature from the utilities used in online learning, $u(a)$, to the analogous action-values used in RL, $q^{\pi}(a)$. 

We start by unifying notations, specifically reformulating the RD dynamics~\cref{eq:rd} in RL terms as
\begin{align}
    \dot \pi(a) = \pi(a)[q^{\bec{\policy}}(a) - v^{\bec{\policy}}].
    \label{eq:rd_rl_terms}
\end{align}
As RD aligns with FoReL with entropy regularization, we can further write the RD logit dynamics using~\cref{eq:forel_logits} as
\begin{align}
    \dot y(a) = q^{\bec{\policy}}(a) - v^{\bec{\policy}},
    \label{eq:nerd_continuous}
\end{align}
where $v^{\bec{\policy}}$ is the variance-reducing baseline~\citep{Sutton18}.
Let $y(a;\btheta_{t})$ denote the logits parameterized by $\btheta_{t}$.
A natural way to derive a parametric update rule is to compute the Euler discretization\footnote{Given a tabular softmax policy, this definition matches the standard discrete-time RD. See \cref{sec:standard_dt_rd}.}
of~\cref{eq:nerd_continuous},
\begin{align}
    y_t(a) \as y(a;\btheta_{t - 1}) + \eta_t \big( q^{\bpi_t}(a) - v^{\bpi_t} \big),
    \label{eq:euler_rd}
\end{align}
and consider $y_t(a)$ a fixed \defword{target value} that the parameterized logits $y(a;\btheta_{t - 1})$ are adjusted toward.  
Namely, one can update $\btheta$ to minimize a choice of metric $d(\cdot,\cdot)$,
\begin{align}
    \btheta_t &= \btheta_{t - 1} - \tsum_a \nabla_{\btheta} d(y_t(a),y(a;\btheta_{t - 1})).
\end{align}
In particular, minimizing the Euclidean distance yields,
\begin{align}
    \btheta_t 
        &= \btheta_{t - 1} - 
            \tsum_a \nabla_{\btheta} \frac{1}{2} \norm{y_t(a) - y(a;\btheta_{t - 1})}^2\\
    &= \btheta_{t - 1} + \tsum_a (y_t(a) - y(a;\btheta_{t - 1})) \nabla_{\btheta} y(a;\btheta_{t - 1}) \\
    & \stackrel{\cref{eq:euler_rd}}{=} \btheta_{t - 1} + \eta_t \tsum_a \nabla_{\btheta} y(a;\btheta_{t - 1}) \big( q^{\bpi}(a) - v^{\bpi} \big),
    \label{eq:nerd_discrete}
\end{align}
which we later prove has a rigorous connection to Hedge and, thus, inherits no-regret guarantees that are useful in nonstationary settings such as games.
Update rule \cref{eq:nerd_discrete} applies to all differentiable policy parameterizations.
However, as our core experiments use neural networks, we henceforth refer to the update rule \cref{eq:nerd_discrete} as \neurdfull (\neurd), with pseudocode shown in \cref{alg:main}.
\neurd effectively corresponds to a `one-line fix' of SPG, in that replacing \cref{alg:main} line~\ref{eq:one_line_change} with \cref{eq:pg_discrete} yields SPG. 

\begin{algorithm}[t]
    \caption{\neurdfull{} (\neurd)}
    \label{alg:main}
    \begin{algorithmic}[1]
        \algrenewcommand\algorithmicindent{1.0em}%
        \State{Initialize policy weights $\btheta_0$ and critic weights $\bw_0$.}
        \For{$t \in \{1,2,\ldots\}$}
            \State{$\bpi_{t-1}(\btheta_{t-1}) \gets \softmaxProjection(\logits(\btheta_{t-1}))$}
            \ForAll{$\tau \in \texttt{SampleTrajectories}(\bpi_{t-1})$} 
                \For{$s, a \in \tau$}\Comment{{\sc policy evaluation}}
                    \State{$R \gets \texttt{Return}(s, \tau, \gamma)$}
                    \State{$\bw_{t} \gets \texttt{UpdateCritic}(\bw_{t-1},s,a,R)$}
                \EndFor
                \For{$s \in \tau$}\Comment{{\sc policy improvement}} 
                    \State{${v}(s;\bw_t) \gets \sum_{a'}\pi(s,\!a'\!;\!\btheta_{t-1})\,{q}_{t}(s,a'\!;\!\bw_{t})$}
                    \State{$\btheta_t\!\gets\!\btheta_{t - 1} + \eta_t\!\sum_{a'}\!\nabla_{\btheta} y(s,a'\!;\!\btheta_{t-1}) \big( {q}_{t}(s,a'\!;\!\bw_{t}) - {v}(s;\bw_t) \big)$
                    \label{eq:one_line_change}
                    }
                \EndFor
            \EndFor
        \EndFor
        \vspace{-10pt}
    \end{algorithmic}
\end{algorithm}

\subsection{Properties and Unifying Theorems}\label{sec:properties_and_unifying_theorems}

Overall, \neurd is not only practical to use as it involves a simple modification of SPG with no added computational expense, but also benefits from rigorous links to algorithms with no-regret guarantees, as shown in this section. 
All proofs are in the appendix.

We first state a well-known relationship\footnote{This relationship also holds for the continuous-time variants as well~\citep{Hofbauer09Time}.} between Hedge and replicator dynamics~\citep{Rustichini99,Kleinberg09Multiplicative,Warmuth16Blessing}:
\begin{statement}
\label{stmt:tabular_neurd_is_hedge}
The following are equivalent: a) Hedge~\citep{Freund97,Littlestone94}, b) discrete-time RD~\citep{Cressman03}, and c) single state, all-actions, tabular \neurd.
\label{thm:tabular_nerd_is_hedge}
\end{statement}
We now extend this equivalence to the sequential decision-making setting, providing convergence guarantees in the tabular setting. We first need a few definitions; see~\citet{srinivasan2018actor} for reference.
First, for any given joint policy, $\bpi = \set{\bpi_{i}}_{i=1}^N$, define the reach probability $\rho^{\bpi}(h) = \rho_i^{\bpi}(h) \rho_{-i}^{\bpi}(h)$ as a product of probabilities of all agents' policies along the history of actions $h$ (including chance/nature). Reach probabilities can be split up into agent $i$'s contribution, $\rho^{\bpi}_i$, and those of the other agents, $\rho^{\bpi}_{-i}$. We make the standard assumption of \defword{perfect recall}, so no agent ever forgets any information they have observed along $h$; an \defword{information state}, $s$, is a set of histories consistent with the observations of the agent to act at $s$. Thus, agent $i$'s policy must be defined on information states, $\bpi_i(s) \in  \simplex^{\abs{\Actions(s)}}$, where $\Actions(s)$ is the set of actions available at information state $s$, but we can overload this notation to say that agent $i$'s policy at history $h \in s$ is $\bpi_i(h) = \bpi_i(s)$.

Due to perfect recall, if agent $i$ is to act at $h$, $\rho^{\bpi}_i(h)$ is the same for all $h \in s$, so we refer to it as $\rho_i^{\bpi}(s)$. 
With $\beta_{-i}(\bpi, s) = \sum_{h \in s}\rho_{-i}^{\bpi}(h)$ as the normalizing denominator, we define:
\[
q^{\bpi}_i(s, a) = \frac{1}{\beta_{-i}(\bpi, s)} \sum_{h \in s} \rho_{-i}^{\bpi}(h) q^{\bpi}_i(h, a),
\] where $q^{\bpi}_i(h, a)$ is the expected value of playing $a$ at history $h$ and following $\bpi$ afterward. Let $v^{\bpi}_i(s) = \sum_{a \in \Actions} \pi_i(a | s) q^{\bpi}_i(s, a)$ be the expected value at information state $s$ for agent $i$.

We can now state the following corollaries to Statement \ref{stmt:tabular_neurd_is_hedge} that give regret and Nash equilibrium approximation bounds in the tabular case:
\begin{restatable}[]{corollary}{tabularNerdSequentialThm}
\label{thm:tabular_nerd_sequential_regret}
Consider a sequential decision making task with finite length histories and $N$-agents. Assume that agent $i$ acts according to a softmax tabular policy, $\bpi_{i,t}(s) \propto \exp(\logits_{i,t}(s))$, where $\logits_{i,t}(s) \in \reals^{\abs{\Actions(s)}}$ is a vector of logits for the actions available at information state $s$, and the other agents act arbitrarily. 

Using \neurd as the local learning algorithm in \defword{counterfactual regret minimization (CFR)}~\citep{Zinkevich08} results in the following local logit updates for agent $i$ and all $s, a$:
\[
y_{i,t}(s,a)
= y_{i,t-1}(s,a) + \eta_t(s) \beta_{-i}(\bpi_{t - 1}, s) \subex{q^{\bpi_{t - 1}}_i(s, a) - v^{\bpi_{t-1}}_i(s)},
\]
where $\beta_{-i}(\bpi_{t - 1}, s) \subex{q^{\bpi_{t-1}}_i(s, a) - v^{\bpi_{t-1}}_i(s)}$ are counterfactual regrets, and $\eta_t(s)$ is the stepsize for information state $s$ at time $t$. These updates and the resulting local policies are identical to those of Hedge by Statement \ref{stmt:tabular_neurd_is_hedge}. Therefore, CFR(\neurd) is equivalent to CFR(Hedge)~\citep{Zinkevich08, Brown17Dynamic}.

By the CFR Theorem~\citep{Zinkevich08} and the Hedge regret bound~\citep{Freund97, cesa2006prediction}, when the time-constant learning rate
$\eta_{t}(s) = \sqrt{{2 \ln(\abs{\Actions(s)})}{T}^{-1}}$ is used at each information state,
a CFR(\neurd) agent has its regret with respect to any fixed policy after $T$-updates, $R_{i, T}$, 
upper-bounded by $R_{i, T} \leq |\States_i| \Delta_{u} \sqrt{ 2 \ln \numActions T }$,
where $\States_i$ is the set of information states for agent $i$, $\Delta_{u} = \max_{z,z'}|u(z) - u(z')|$ for any two terminal histories, $z, z'$, and
$\numActions$ is the maximum number of actions at any of agent $i$'s information states.
\end{restatable}

\begin{corollary}
\label{cor:tabular_nerd_nash_approx}
As stated in \cref{sec:online-learning}, \cref{thm:tabular_nerd_sequential_regret} implies that in a two-player, zero-sum game, if each player $i$ employs CFR(NeuRD), the policy generated from their average sequence weights, $\bar{\bpi}_{i, T}(s) \propto  \frac{1}{T} \sum_{t = 1}^T \rho^{\bpi_t}_i(s) \bpi_{i, t}(s)$, converges to an $\epsilon$-Nash equilibrium, with $\epsilon$ upper-bounded by
the maximum $R_{i, T}$ across players by the folk theorem~\citep{Blum07}.
\end{corollary}

\cref{thm:tabular_nerd_sequential_regret} and \cref{cor:tabular_nerd_nash_approx} show that \neurd can be used to solve a broad class of problems where SPG may fail.
Note that in the function approximation case, where we use $y(s, a; \btheta_t)$ to approximate the tabular logit for information state $s$ and action $a$, one can still establish regret guarantees~\citep{dorazio2019alternative}.

We next formalize the connection between RD and PG, expanding beyond the scope of prior works that have considered only the links between EGT and value-iteration based algorithms~\citep{Tuyls03,kaisers2010frequency}.
\begin{restatable}[]{thm}{rdToPgThm}
    SPG is a policy-level Euler discretization approximation of continuous-time RD (i.e., computing $\bpi_{t+1}$ using $\bpi_{t}$), under a KL-divergence minimization criterion.
    \label{thm:rd_to_pg}
\end{restatable}

Next we establish a formal link between \neurd and Natural Policy Gradient (NPG)~\citep{kakade2002natural}.
\begin{restatable}[]{thm}{nerdToNaturalPgThm}
    The \neurd update rule \cref{eq:nerd_discrete} corresponds to a naturalized policy gradient rule, in the sense that \neurd applies a natural gradient only at the policy output level of softmax function over logits, and uses the standard gradient otherwise.
    \label{thm:nerd_to_natural_pg}
\end{restatable}
Unlike NPG, \neurd does not require computation of the inverse Fisher information matrix, which is especially expensive when, e.g., the policy is parameterized by a large-scale neural network~\citep{martens2015optimizing}.

\begin{remark*}
As, on average, the logits $\logits$ get incremented by the advantage, they may diverge to $\pm \infty$. 
To avoid numerical issues, one can stop increasing/decreasing the logits if the logit-gap exceeds a threshold. 
We apply this by using a clipping gradient, $\widehat{\nabla}_{\btheta}$, as follows:
\begin{align}
  \widehat{\nabla}_{\btheta}(z(\btheta), \eta, \beta) \as [\eta\nabla_{\btheta}z(\btheta)] \indic{z\big(\btheta + \eta\nabla_{\btheta}z(\btheta)\big) \in [-\beta,\beta]},
\end{align}
where $\indic{\cdot}$ is the indicator function, $\eta>0$ is a learning rate, and $\beta \in \reals$ controls the allowable logits gap.
This yields the update,
\begin{equation}
    \btheta_t = \btheta_{t - 1} + \widehat{\nabla}_{\btheta}\big(y(s,a;\btheta_{t-1})(q^{\bpi}(s,a) - v^{\bpi}(s)), \eta_t, \beta\big).
    \label{eq:nerd_sampled}
\end{equation}
Note that logits thresholding is not problematic at the policy representation level, since actions can have a probability arbitrarily close to $0$ or $1$ given a large enough $\beta$.
\end{remark*}

\begin{figure}[t]
    \centering
    \begin{subfigure}[t]{0.48\linewidth}
        \centering
        \includegraphics[height=2cm,trim={0 3mm 0 2mm},clip]{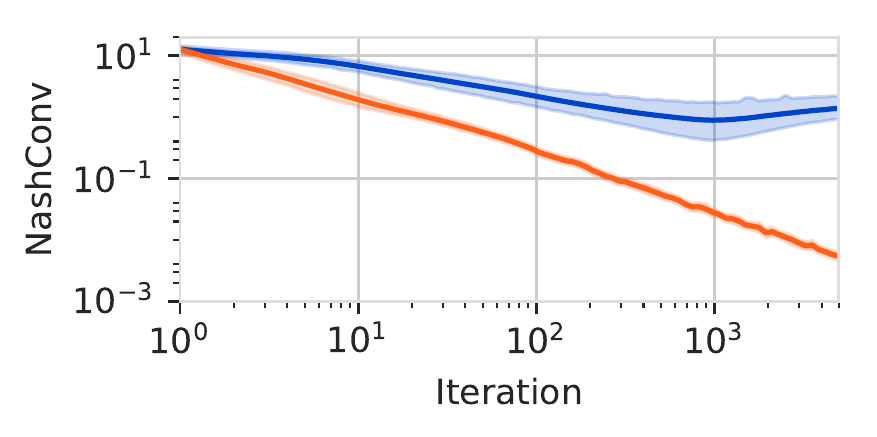}\vspace{-3pt}\caption{Biased RPS}
        \label{fig:nerd_vs_pg_fixed_rps}
    \end{subfigure}%
    \hfill%
    \begin{subfigure}[t]{0.48\linewidth}
        \centering
        \includegraphics[height=2cm,trim={0 3mm 0 2mm},clip]{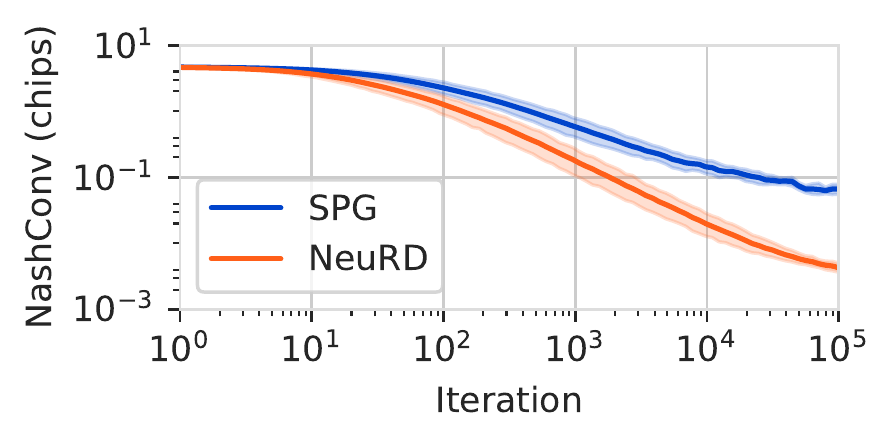}\vspace{-3pt}\caption{Leduc Poker}
        \label{fig:nerd_vs_pg_allaction_leduc}
    \end{subfigure}
    \vspace{-5pt}
    \caption{\subref{fig:nerd_vs_pg_fixed_rps} \NashConv of the average \neurd and SPG policies in biased RPS. \subref{fig:nerd_vs_pg_allaction_leduc} \NashConv{} of the sequence-probability average policies of tabular, all-actions, counterfactual value \neurd and SPG in two-player Leduc Poker.
    }
    \label{fig:tabular_leduc}\vspace{-10pt}
\end{figure}

\section{Evaluation}\label{sec:evaluation}

We conduct a series of evaluations demonstrating the effectiveness of \neurd when learning in nonstationary settings such as NFGs, standard imperfect information benchmarks, and variants of each with added reward nonstationarity.
As \neurd involves only a simple modification of the SPG update rule to improve adaptivity, we focus our comparisons against SPG as a baseline, noting that additional benefits can be analogously gained by combining \neurd with more intricate techniques that improve SPG (e.g., variance reduction, improved exploration, or off-policy learning).

We consider several domains.
\textbf{Rock--Paper--Scissors (RPS)} is a canonical NFG involving two players. 
In \textbf{Kuhn Poker}, each player starts with 2 chips, antes 1 chip to play, receives a face-down card from a deck of $K+1$ such that one card remains hidden, and either bets (raise or call) or folds until all players are in (contributed equally to the pot) or out (folded).
Amongst those that are in, the player with the highest-ranked card wins the pot.
In \textbf{Leduc Poker}~\citep{Southey05}, players instead have limitless chips, one initial private card, and ante 1 chip to play. 
Bets are limited to 2 and 4 chips, respectively, in the first and second round, with two raises maximum in each round.
A public card is revealed before the second round so at a showdown, the player whose card pairs the public card or has the highest-ranked card wins the pot.
In \textbf{Goofspiel}, players try to obtain point cards by bidding simultaneously.
We use an imperfect information variant with 5 cards where bid cards are not revealed~\citep{lanctot13phdthesis}. 
The number of information states for the Kuhn, Leduc, and Goofspiel variants evaluated is 12, 936, and 2124, respectively;
we use the OpenSpiel~\citep{lanctot2019openspiel} implementations of these three games in our experiments. 

We first show that the differences between \neurd and SPG detailed in \cref{sec:properties_and_unifying_theorems} are more than theoretical.
Consider the \NashConv of the time-average \neurd and SPG tabular policies in the game of RPS, shown in \cref{fig:nerd_vs_pg_fixed_rps}. 
Note that by construction, \neurd and RD are equivalent in this tabular, single-state setting. 
\neurd not only converges towards the Nash equilibrium faster, but SPG eventually plateaus.
The continuous time dynamics of \neurd~\cref{eq:nerd_continuous} and SPG~\cref{eq:qpg} are integrated over time with step size $\Delta t=0.1$ (equals 1 iteration). The figure shows the mean \NashConv~\cref{eq:nashconv} of 100 trajectories starting from initial conditions sampled uniformly from the policy simplex. The shaded area corresponds to the $95\%$ confidence interval computed with bootstrapping from 1000 samples.

\begin{figure}[t]
    \centering
    \includegraphics[height=2cm,trim={0 3mm 0 2mm},clip]{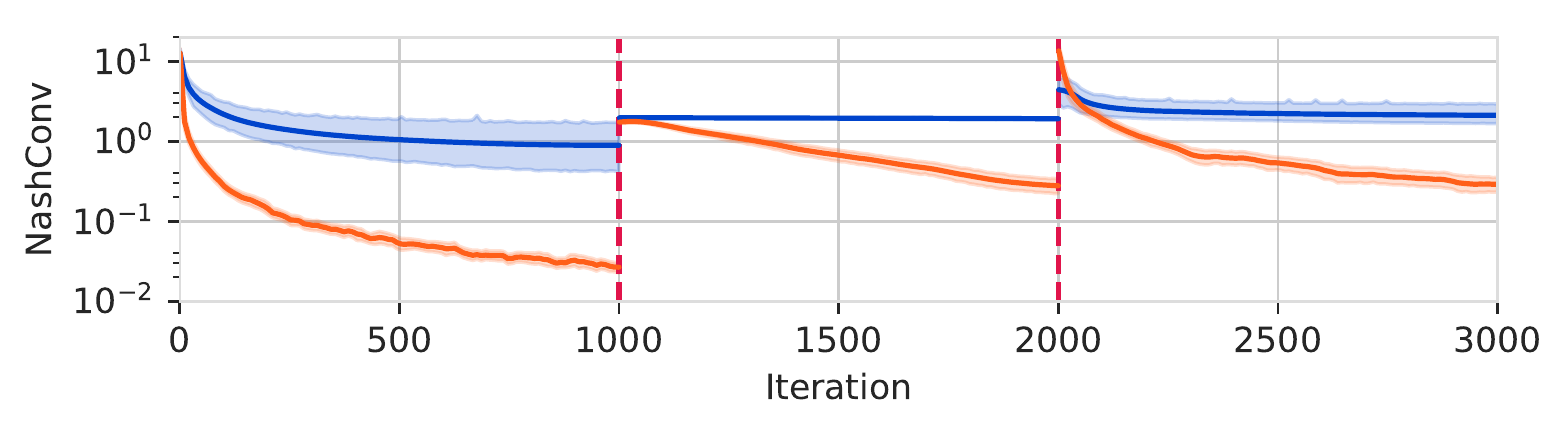}
    \vspace{-7pt}
    \caption{
    Time-average policy \NashConv in nonstationary RPS, with the game phases separated by vertical red dashes.
    }
    \label{fig:nonstationary_rps}\vspace{-10pt}
\end{figure}

Consider next a more complex imperfect information setting, where \cref{fig:tabular_leduc} shows that tabular, all-actions, counterfactual value \neurd\footnote{This can be seen as counterfactual regret minimization (CFR)~\citep{Zinkevich08} with Hedge~\citep{Brown17Dynamic}.}
more quickly and more closely approximates a Nash equilibrium in two-player Leduc Poker than tabular, all-actions, counterfactual value PG. 
In every iteration, each information-state policy over the entire game was updated for both players in an alternating fashion, i.e., the first player's policy was updated, then the second player's~\citep[Section 4.3.6]{Burch2017}~\citep{Burch2019}. The only difference between the \neurd and SPG algorithms in this case is the information-state logit update rule, and the only difference here--as described by~\cref{eq:tabular_all_actions_pg_update}--is that SPG scales the \neurd update by the current policy. The performance displayed is that of the sequence probability time-average policy for both algorithms. The set of constant step sizes tried were the same for both algorithms:
$
\stepSize \in \set{0.5, 0.9, 1, 1.5, 2, 2.5, 3, 3.5, 4}
$.
The shaded area corresponds to the $95\%$ interval that would result from a uniform sampling of the step size from this set.

\begingroup
\setlength{\columnsep}{10pt}
\setlength{\intextsep}{2pt}
\setlength{\abovecaptionskip}{0pt}
\setlength{\belowcaptionskip}{0pt}
\begin{wraptable}{r}{0.35\linewidth} 
    \centering
    \begin{tabular}{c|ccc}
     & R & P & S \\ \hline
     R & \hphantom{-}0  & -1            & \hphantom{-}$\nu$\\
     P & \hphantom{-}1  & \hphantom{-}0 & -1 \\
     S & -$\nu$         & \hphantom{-}1 & \hphantom{-}0
    \end{tabular}
    \caption{RPS payoffs.}\label{eq:biased-rps}
\end{wraptable}
We next consider modifications of our domains wherein reward functions change at specific intervals during learning, compounding the usual nonstationarities in games.
Specifically, we consider games with three phases, wherein learning commences under a particular reward function, after which it switches to a different function in each phase while learning continues \emph{without} the policies being reset.
In biased RPS, each phase corresponds to a particular choice of the parameter $\nu$ in \cref{eq:biased-rps}.

\endgroup

Payoffs are shown for the first player only.
We set $\nu$ to 20, 0, and 20, respectively, for the three phases, with payoff switches happening every 1000 iterations.
This effectively biases the Nash towards one of the simplex corners (see \cref{fig:biased_rps_rd}), then to the center (\cref{fig:rps_rd}), then again towards the corner.
\Cref{fig:nonstationary_rps} plots the \NashConv of \neurd and SPG with respect to the Nash for that particular phase. 
Despite the changing payoffs, the \neurd \NashConv decreases in each of the phases, while SPG plateaus.
We use the same setup detailed above for~\cref{fig:nerd_vs_pg_fixed_rps} for the nonstationary case.

We next consider imperfect information games, with the reward function being iteratively negated in each game phase for added nonstationarity, and policies parameterized using neural networks.
Due to the complexity of maintaining a time-average neural network policy to ensure no-regret, we instead use entropy regularization of the form introduced by \citet{perolat2020poincar} to induce realtime policy convergence towards the Nash. 
Specifically, \citet{perolat2020poincar} show that the last iterate policy of FoReL will converge to the Nash equilibrium in zero-sum two-player imperfect information games when an additional entropy cost is used.
We note that the empirical evaluation of QPG in \citet{srinivasan2018actor} corresponds to SPG in our work in that both are all-actions policy gradient with a softmax projection. However, the form of entropy regularization in~\citet{srinivasan2018actor} uses the common approach of adding an entropy bonus to the policy loss, whereas here we use entropy regularization applied to the action-values $q^{\bpi}$~\citep{perolat2020poincar}.

\begin{figure*}[t]
    \centering
    \newcommand{\figHeight}{2.09cm}
    \begin{subfigure}[c]{0.17\linewidth}
        \centering
        \includegraphics[height=\figHeight]{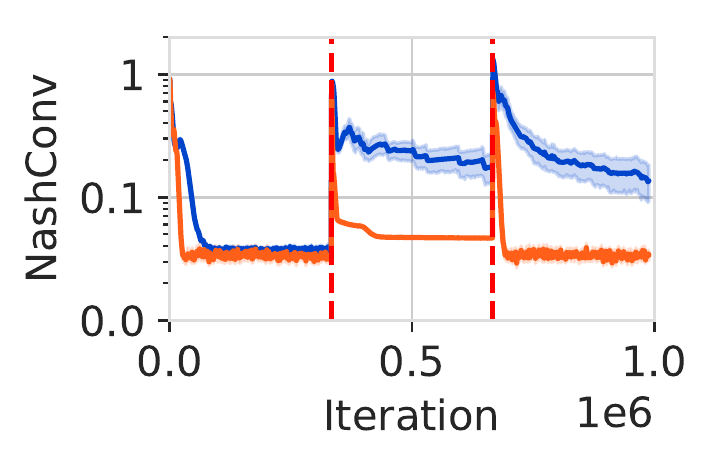}
        \caption{Kuhn Poker}
        \label{fig:nonstationary_kuhn_poker_bestparams__mainpaper_nashconv}
    \end{subfigure}%
    \begin{subfigure}[c]{0.17\linewidth}
        \centering
        \includegraphics[height=\figHeight]{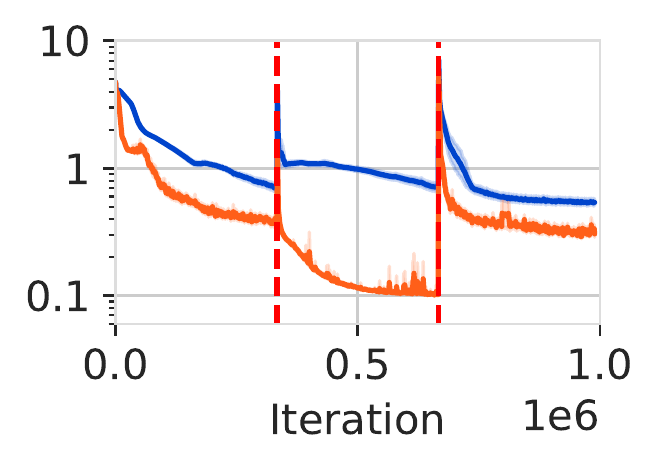}
        \caption{Leduc Poker}
        \label{fig:nonstationary_leduc_poker_bestparams__mainpaper_nashconv}
    \end{subfigure}%
    \begin{subfigure}[c]{0.17\linewidth}
        \centering
        \includegraphics[height=\figHeight]{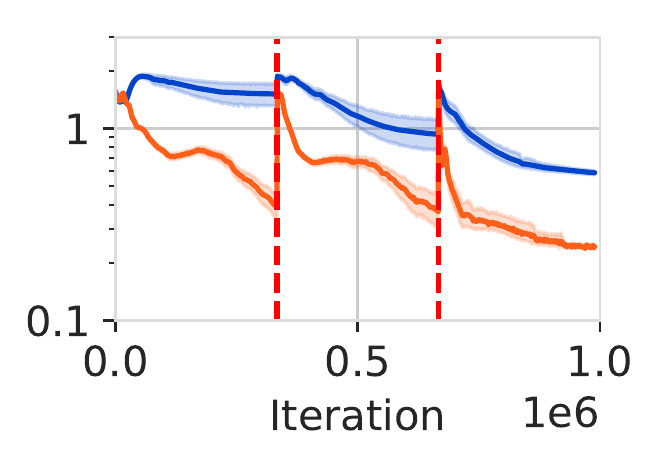}
        \caption{Goofspiel}
        \label{fig:nonstationary_goofspiel_large_bestparams__mainpaper_nashconv}
    \end{subfigure}%
    \rulesep
    \begin{subfigure}[c]{0.15\linewidth}
        \includegraphics[height=\figHeight]{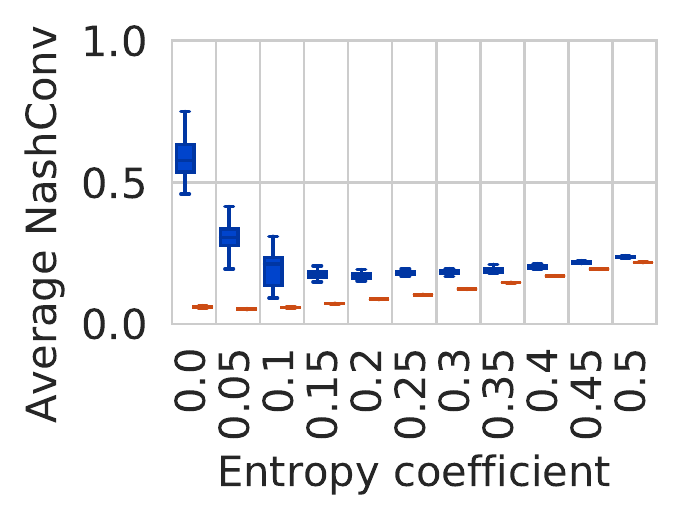}
        \caption{Kuhn Poker}
        \label{fig:exploitability_merged_AUC_kuhn_poker_all_tranche_boxplot_nerd}
    \end{subfigure}%
    \hfill%
    \begin{subfigure}[c]{0.15\linewidth}
        \includegraphics[height=\figHeight]{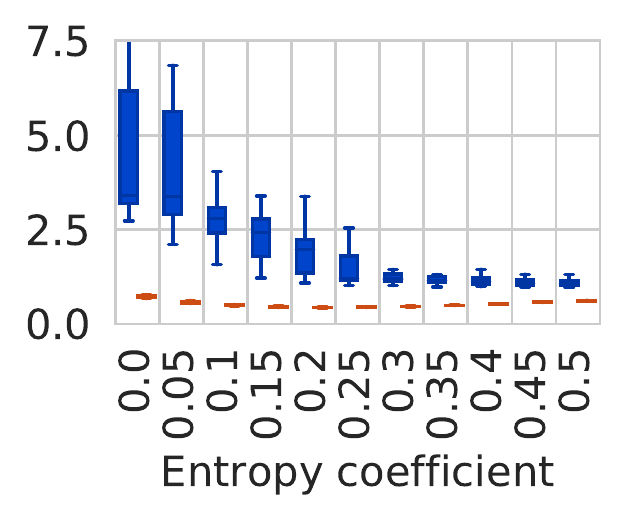}
        \caption{Leduc Poker}
        \label{fig:exploitability_merged_AUC_leduc_poker_all_tranche_boxplot_nerd}
    \end{subfigure}%
    \hfill%
    \begin{subfigure}[c]{0.15\linewidth}
        \includegraphics[height=\figHeight]{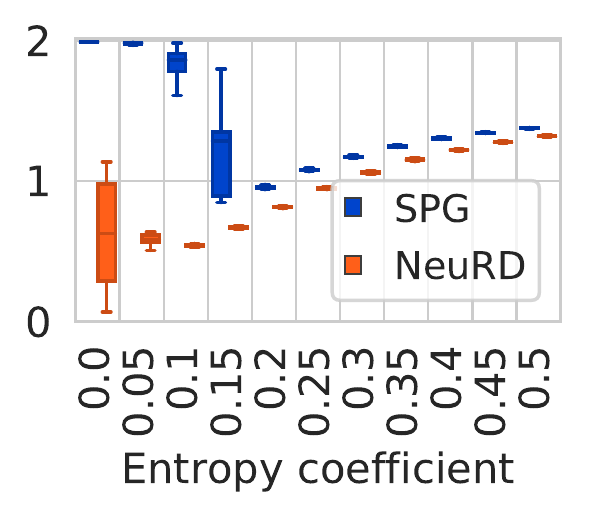}
        \caption{Goofspiel}
        \label{fig:exploitability_merged_AUC_goofspiel_large_all_tranche_boxplot_nerd}
    \end{subfigure}%
    \vspace{-10pt}
    \caption{Comparison of \neurd and SPG \NashConv (left) and average \NashConv over all iterations (right), for each game.}
    \label{fig:openspiel_results}
    \vspace{-7pt}
\end{figure*}

For each game in \cref{fig:openspiel_results}, we randomly initialize a policy parameterized by a two-layer neural network (128 hidden units). Results are reported for 40 random neural network initializations for both \neurd and PG.
We update the policy once every 4 updates of the Q-function. 
The batch size of the policy update is 256 trajectories, with trajectory lengths of 5, 8, and 8 for Kuhn Poker, Leduc Poker, and Goofspiel. The Q-function update batch size is 4 trajectories (same lengths). A learning rate of 0.002 was used for policy updates, and 0.01 for Q-value function updates.
Reward function negation occurs every $0.33\mathrm{e}6$ learning iterations (with the three reward function phases separated by the vertical red stripes in plots, where applicable).
Upon conclusion of each learning phase, policies are not reset; instead, learning continues given the latest policy (for each of the 40 trials).

\Cref{fig:nonstationary_kuhn_poker_bestparams__mainpaper_nashconv,fig:nonstationary_leduc_poker_bestparams__mainpaper_nashconv,fig:nonstationary_goofspiel_large_bestparams__mainpaper_nashconv} illustrate the \NashConv for \neurd and SPG in all imperfect information games considered, with an entropy regularization sweep conducted for each algorithm to yield the lowest \emph{final} \NashConv.
Notably, \neurd converges faster than SPG in all three domains, and yields the lowest final \NashConv.
To analyze the role of entropy regularization on the convergence {rate}, we next consider the average \NashConv over \emph{all} training iterations and phases of each game.
\Cref{fig:exploitability_merged_AUC_kuhn_poker_all_tranche_boxplot_nerd,fig:exploitability_merged_AUC_leduc_poker_all_tranche_boxplot_nerd,fig:exploitability_merged_AUC_goofspiel_large_all_tranche_boxplot_nerd} plot the average \NashConv for each game, across all considered entropy regularization levels.
Notably, \neurd consistently has lower average \NashConv and is less sensitive to the entropy regularization level, in contrast to PG.

Overall, the key takeaway of these experiments is SPG's weakness when learning in nonstationary domains.
Critically, this weakness is effectively addressed through a simple `one-line change' in using the \neurd update rule \cref{eq:nerd_discrete} in lieu of the standard policy gradients update, as illustrated in \cref{alg:main}.

\section{Discussion}

We established here unifying, rigorous links between replicator dynamics, policy gradient methods, and online learning.
We began by demonstrating that commonly-used SPG methods face major adaptivity issues, even in the simplest of nonstationary domains.
The insights gained led to development of a novel algorithm, \neurd, which generalizes the no-regret Hedge algorithm and RD to utilize function approximation. \neurd is thus theoretically grounded and adaptive while still enjoying the practical benefits of parametric SPG methods. 
The key advantage of \neurd is that it corresponds to a simple `one-line fix' of standard PG algorithms. 

\neurd was established to have a number of theoretical properties, including regret guarantees in the sequential setting, in addition to concrete theoretical links to SPG and Natural PG.
We empirically showed \neurd to significantly outperform SPG in numerous highly nonstationary and adversarial settings, ranging from tabular to imperfect information games. 
Moreover, in contrast to existing sequence-form dynamics \citep{Gatti13Efficient,Lanctot14Further,Panozzo14SFQ,Gatti16LogitDynamics}, use of counterfactual values in \neurd enables representation of policies in behavioral form, as is standard in RL.
As a result, it is straight-forward to conduct sampling and function approximation via \neurd.

While \neurd represents an important extension of classical learning dynamics to utilize function approximation, Hedge and RD are also instances of the more general FoReL framework that applies to arbitrary convex decision sets and problem geometries expressed through a regularization function.
This connection suggests that \neurd could perhaps likewise be generalized to convex decision sets and various parametric forms, which would allow a general FoReL-like method to take advantage of functional representations.
Moreover, as \neurd involves a very simple modification of the SPG update rule, a promising avenue for future work is to investigate \neurd-based extensions of standard PG-based methods (e.g., A3C~\citep{mnih2016asynchronous}, DDPG~\citep{lillicrap2015continuous}, and MADDPG~\citep{lowe2017multi}), opening the door to a large class of new and potentially performative algorithms. 
Additionally, it seems sensible to investigate nonstationary single-agent RL tasks such as intrinsic motivation-based exploration~\citep{graves2017automated}.

\begin{appendices}
\crefalias{section}{appsec} 
\crefalias{subsection}{appsec} 
\crefalias{appendixproof}{appsec} 

\renewcommand\thefigure{\thesection.\arabic{figure}}    
\renewcommand\theparagraph{\thesection.\arabic{paragraph}}    

\renewcommand{\appendixproof}[1]{
    \par
    \refstepcounter{apcounter}
    \noindent
    \textbf{\thesection.\theapcounter}\space\space
    \textbf{#1}\space\space
    }

\section*{Appendices}\label{sec:proofs}
\addtocounter{section}{1}

\appendixproof{Single state, tabular SPG update.}\label{sec:tabular_pg_update}
Here, we fully derive the single state, tabular SPG update.
On round $t$, SPG updates its logits as
$
    \bec{\logit}_{t} 
        \as \bec{\logit}_{t - 1} + \stepSize_t \grad_{\bec{\logit}_{t - 1}} \bec{\policy}_{t}\cdot\bec{\utility}_t.
$
As there is no action or state sampling in this setting, shifting all the payoffs by the expected value $\bar{u}$ has no impact on the policy, so this term is omitted here. 
Noting that $\nicefrac{\partial \policy_t(\action')}{\partial \logit_{t - 1}(\action)} = \policy_t(\action') \subblock{ \indicator{\action' = \action} - \policy_t(\action) }$~\citep[Section 2.8]{Sutton18}, we observe that the update direction, $\grad_{\bec{\logit}_{t - 1}} \bec{\policy}_t \cdot \bec{\utility}_t$, is actually the instantaneous regret scaled by $\bec{\policy}_t$:
\begin{align}
    \frac{\partial \bec{\policy}_t}{\partial \logit_{t - 1}(\action)} \cdot \bec{\utility}_t
        &= \sum_{\action'}
            \frac{\partial \policy_t(\action')}{\partial \logit_{t - 1}(\action)} \utility_t(\action')\\
        &= \tsum_{\action'}
            \policy_t(\action')
            \subblock{ \indicator{\action' = \action} - \policy_t(\action) }
            \utility_t(\action')\\
        &= \policy_t(\action) \subblock{ 1 - \policy_t(\action) } \utility_t(\action)
            - \tsum_{\action' \neq \action}
                \policy_t(\action') \policy_t(\action) \utility_t(\action')\\
        &= \policy_t(\action) \big[
            \utility_t(\action) - \policy_t(\action) \utility_t(\action)
            - \tsum_{\action' \neq \action}
                \policy_t(\action') \utility_t(\action') \big]\\
        &= \policy_t(\action) \big[
            \utility_t(\action) 
            - \tsum_{\action'} \policy_t(\action') \utility_t(\action')\big].
\end{align}
Therefore, the concrete update is:
\begin{align}
    \logit_t(\action) 
        &= \logit_{t - 1}(\action) + \stepSize_t \policy_t(\action) \big[
        \utility_t(\action)
            - \tsum_{\action'} \policy_t(\action') \utility_t(\action')
        \big]\\
        &= \logit_{t - 1}(\action) + \stepSize_t \policy_t(\action) \big[
        \utility_t(\action)
            - \bar{u}_t
        \big].
\end{align}

\appendixproof{Proof of \cref{thm:tabular_nerd_is_hedge}.}
    In the single state, tabular case, $\grad_{\btheta} \logit(\action ; \btheta_t)$ is the identity matrix, so unrolling the \neurd update~\cref{eq:nerd_discrete} across $T - 1$ rounds, we see that the \neurd policy is
    \begin{align}
        \policy_T
            &= \softmaxProjection \subex{ \textstyle \tsum_{t = 1}^{T - 1} \stepSize_t \subex{ \bm{\utility}_t - \bm{\utility}_t \cdot \policy_t }}
            = \softmaxProjection \subex{\textstyle  \tsum_{t = 1}^{T - 1} \stepSize_t \bm{\utility}_t },
        \label{eq:tabular_nerd_policy}
    \end{align}
    as $\softmaxProjection$ is shift invariant. As~\cref{eq:tabular_nerd_policy} is identical to~\cref{eq:hedge}, \neurd and Hedge use the same policy on every round, thus are equivalent here.
\appendixproof{Proof of \cref{thm:rd_to_pg}.}\label{sec:rd_to_pg}
    An Euler discretization of RD at the policy level is:
    $ \bpi_{t + 1} \as  \bpi_t \odot \Big[ \ones + \eta_t \big( \bm{q} - v^{\pi_t} \ones \big) \Big]$.
    Note that while $\sum_a \pi_{t+1}(a) = 1$, this Euler-discretized update may still be outside the simplex; however, $\bpi_{t+1}$ merely provides a target for our parameterized policy $\bpi_{\btheta_t}$ update, which is subsequently reprojected back to the simplex via $\softmaxProjection(\cdot)$. 
    
    Now if we consider parameterized policies $\bpi_{t} \approx \bpi_{\btheta_t}$, and our goal is to define dynamics on $\btheta_t$ that captures those of RD, a natural way consists in updating $\btheta_{t}$ in order to make $\bpi_{\btheta_{t}}$ move towards $\bpi_{t + 1}$, for example in the sense of minimizing their KL divergence, $KL(\bm{p}, \bm{q}) \as \bm{p} \cdot \log \bm{p} - \bm{p} \cdot \log \bm{q}, \bm{p}, \bm{q} \in \reals^{+, n}, n > 0$.
    
    Of course, the KL divergence is defined only when both inputs are in the positive orthant, $\reals^{+, n}$, so in order to measure the divergence from $\bpi_{t + 1}$, which may have negative values, we need to define a KL-like divergence. Fortunately, since the $\bm{p} \cdot \log \bm{p}$ is inconsequential from an optimization perspective and this is the only term that requires $\bm{p} > 0$, a natural modification of the KL divergence to allow for negative values in its first argument is to drop this term entirely, resulting in $\tilde{KL}(\bm{p}, \bm{q}) \as -\bm{p} \cdot \log \bm{q}, \bm{p} \in \reals^n, \bm{q} \in \reals^{+, n}$.
    
    The gradient-descent step on the $\tilde{KL}(\bpi_{t + 1}, \bpi_{\btheta_t})$ objective is:
    \begin{align}
        \btheta_{t+1} 
            &= \btheta_{t} - \nabla_{\btheta} \tilde{KL}(\bpi_{t + 1}, \bpi_{\btheta_{t}}) \\
            &= \btheta_{t} + \tsum_a \pi_{t + 1}(a) \nabla_{\btheta} \log \pi_{\btheta_{t}}(a) \\
            &= \btheta_{t} + 
                \tsum_a 
                    \pi_t(a)\subblock{
                        1  + \eta_t \big( q(a) - v^{\pi_t} \big)
                    } \nabla_{\btheta} \log \pi_{\btheta_{t}}(a). \\
        \shortintertext{Assuming $\pi_t = \pi_{\btheta_{t}}$,}
        \btheta_{t+1} 
            &= \btheta_{t} + \tsum_a \pi_{\btheta_{t}}(a)\Big[1  + \eta_t \big( q(a) - v^{\pi_{\btheta_{t}}} \big) \Big] \nabla_{\btheta} \log \pi_{\btheta_{t}}(a) \\
            &= \btheta_{t} + \tsum_a \Big[1  + \eta_t \big( q(a) - v^{\pi_{\btheta_{t}}} \big) \Big] \nabla_{\btheta} \pi_{\btheta_{t}}(a) \\
            &= \btheta_{t} +  \big(1  - \eta_t v^{\pi_{\btheta_{t}}} \big) \tsum_a   \nabla_{\btheta} \pi_{\btheta_{t}}(a) + \eta_t \tsum_a   q(a) \nabla_{\btheta} \pi_{\btheta_{t}}(a) \\
            &= \btheta_{t} + \eta_t  \nabla_{\btheta} \tsum_a \pi_{\btheta_{t}}(a) q(a)
            = \btheta_{t} + \eta_t  \nabla_{\btheta} v^{\pi_{\btheta_{t}}},
    \end{align}
    which is precisely a policy gradient step.

\appendixproof{Proof of \cref{thm:nerd_to_natural_pg}.}\label{sec:nerd_and_pg}
    Consider a policy $\pi(a)$ defined by a softmax over a set of logits $y(a)$: $\pi \as \softmaxProjection(y)$.
    Define the Fisher information matrix $F$ of the policy $\pi$ with respect to the logits $y$:
    \begin{align}
        F_{a,b} &= \tsum_c \pi(c) (\partial_{y(a)}\log \pi(c))(\partial_{y(b)}\log \pi(c)) \\
         &= \begin{aligned}[t]
            \tsum_c \pi(c) (\partial_{y(a)} y(c) &-\tsum_d \pi(d) \partial_{y(a)} y(d))(\partial_{y(b)} y(c)\\
            &-\tsum_d \pi(d) \partial_{y(b)} y(d))
        \end{aligned}\\
        &= \tsum_c \pi(c) (\delta_{a,c} -\tsum_d \pi(d) \delta_{a,d})(\delta_{b,c} -\tsum_d \pi(d) \delta_{b,d})\\
        &= \pi(b) (\delta_{a,b} -\pi(a)).
    \end{align}
    \begin{align}
    \text{Note that \,\,}
        (F\nabla y)(a) &= \tsum_b F_{a,b} \nabla y(b) \\
        &= \tsum_b \pi(b) (\delta_{a,b}-\pi(a))\nabla y(b) \\
        &= \pi(a) \nabla y(a) - \pi(a) \tsum_b \pi(b) \nabla y(b) = \nabla \pi(a)
    \end{align}
    from the definition of $\pi$. 
    Considering the variables $y$ as parameters of the policy, the natural gradient $\widetilde\nabla_y \pi$ of $\pi$ with respect to $y$ is 
        $\widetilde\nabla_y \pi = F^{-1} (\nabla_y \pi) = F^{-1} (F \nabla_y y) = I.$
    Now assume the logits $y_{\btheta}$ are parameterized by some parameter $\btheta$ (e.g., with a neural network). 
    Let us define the \emph{pseudo-natural gradient} of the probabilities $\pi$ with respect to $\btheta$ as the composition of the natural gradient of $\pi$ with respect to $y$ (i.e., the softmax transformation) and the gradient of $y_{\btheta}$ with respect to $\btheta$, which is
        $\widetilde\nabla_{\btheta} \pi =  (\nabla_{\btheta} y_{\btheta})(\widetilde\nabla_y \pi) = \nabla_{\btheta} y_{\btheta}.$
    Thus, we have that a natural policy gradient yields:
    \begin{align}
        \tsum_a \widetilde\nabla_{\btheta} \pi(a) \big( q(a) - v^{\pi}\big) 
        = \tsum_a \nabla_{\btheta} y(a) \big( q(a) - v^{\pi}\big), 
    \end{align}
    which is nothing else than the \neurd update rule in \cref{eq:nerd_discrete}.

\appendixproof{Equivalence to standard discrete-time RD.}\label{sec:standard_dt_rd}
A common way to define discrete-time replicator dynamics is according to the so-called \emph{standard discrete-time replicator dynamic}~\citep{Cressman03}, 
$   \policy_t(\action)
        \as
            \policy_{t - 1}(\action)
            \e{q^{\policy_{t - 1}(\action)}} (\bpi_{t - 1} \cdot \e{\bm{q}^{\policy_{t - 1}}})^{-1}.$
The action values are exponentiated to ensure all the utility values are positive, which is the typical assumption required by this model. Since the policy is a softmax function applied to logits, we can rewrite this dynamic in the tabular case to also recover an equivalent to the \neurd update rule in \cref{eq:nerd_discrete} with $\learningRate = 1$:
\begin{align}
    \policy_{t - 1}(\action)
            \frac{\e{q^{\policy_{t - 1}}(\action)}}{\bpi_{t - 1} \cdot \e{\bm{q}^{\policy_{t - 1}}}}
        &= \frac{\e{y_{t - 1}(\action) + q^{\policy_{t - 1}}(\action)}}{\sum_b \e{y_{t - 1}(b)}} \frac{\sum_b \e{y_{t - 1}(b)}}{\sum_b \e{y_{t - 1}(b) + q^{\policy_{t - 1}}(b)}}\\
        &= \frac{\e{y_{t - 1}(\action) + q^{\policy_{t - 1}}(\action)}}{\sum_b \e{y_{t - 1}(b) + q^{\policy_{t - 1}}(b)}}.
\end{align}
$\policy_t$ is generated from logits, $y_t(\action) = y_{t - 1}(\action) + q^{\policy_{t - 1}}(\action) = \sum_{\tau = 0}^{t - 1} q^{\policy_{\tau}}(\action)$, which only differs from \cref{eq:nerd_discrete} in a constant shift of $v^{\policy_{t - 1}}$ across all actions.
Since the softmax function is shift invariant, the sequence of policies generated from these update rules will be identical.

\end{appendices}

\bibliographystyle{ACM-Reference-Format}
\bibliography{paper}

\end{document}